\documentclass[runningheads]{llncs}
\usepackage{graphicx}
\usepackage{caption}
\usepackage{tikz}
\usepackage{comment}
\usepackage{amsmath,amssymb} 
\usepackage{color, multirow}
\usepackage[ruled,vlined]{algorithm2e}
\usepackage{hyperref}
\usepackage{xcolor,colortbl}
\usepackage[accsupp]{axessibility}  

\usepackage{bbm}

\begin{document}
\pagestyle{headings}
\mainmatter
\def\ECCVSubNumber{7331}  

\title{Autoregressive Uncertainty Modeling for 3D Bounding Box Prediction} 

\titlerunning{Autoregressive 3D Bounding Box}
%
\author{YuXuan Liu\inst{1,2} \and
Nikhil Mishra\inst{1,2} \and
Maximilian Sieb\inst{1} \and
Yide Shentu\inst{1,2} \and
Pieter Abbeel\inst{1, 2} \and
Xi Chen\inst{1}}
\authorrunning{Y. Liu et al.}
%
\institute{Covariant  \and UC Berkeley \\ \email{yuxuanliu@berkeley.edu}}
\maketitle

\begin{abstract}
3D bounding boxes are a widespread intermediate representation in many computer vision applications.
However, predicting them is a challenging task, largely due to partial observability, which motivates the need for a strong sense of uncertainty.
While many recent methods have explored better architectures for consuming sparse and unstructured point cloud data, we hypothesize that there is room for improvement in the modeling of the output distribution and explore how this can be achieved using an autoregressive prediction head.
Additionally, we release a simulated dataset, COB-3D, which highlights new types of ambiguity that arise in real-world robotics applications, where 3D bounding box prediction has largely been underexplored.
We propose methods for leveraging our autoregressive model to make high confidence predictions and meaningful uncertainty measures, achieving strong results on SUN-RGBD, Scannet, KITTI, and our new dataset\footnote{Code and dataset are available at \href{https://bbox.yuxuanliu.com}{bbox.yuxuanliu.com}}.

\end{abstract}

\section{Introduction}

\begin{figure}[t]
    \centering
    \includegraphics[width=\textwidth]{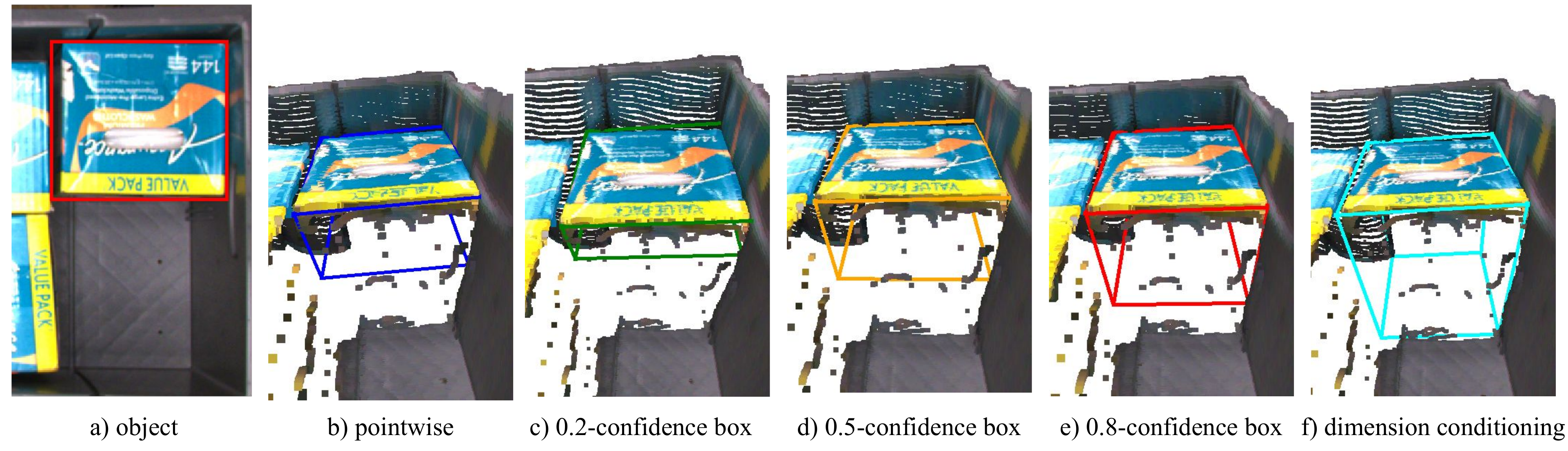}
    \caption{
    a) In this scene from a real-world robotics application, how tall is the object highlighted in red? 
    b) A pointwise model could output only one box prediction with no notion of uncertainty
    c)-e) Predictions from our confidence box method. Notice that the predicted box expands in the direction of uncertainty as we increase the confidence requirement.
    f) Our dimension conditioning method can leverage additional information to make more accurate predictions.
    }
    \label{fig:stacked_boxes}
\end{figure}
Predicting 3D bounding boxes is a core part of the computer vision stack in many real world applications, including autonomous driving, robotics, and augmented reality. 
The inputs to a 3D bounding box predictor usually consist of an RGB image and a point cloud; the latter is typically obtained from a 3D sensor such as LIDAR or stereo depth cameras.
These 3D sensing modalities have their own idiosyncrasies: LIDAR tends to be accurate but very sparse, and stereo depth can be both sparse and noisy.
When combined with the fact that objects are only seen from one perspective, the bounding-box prediction problem is fundamentally underspecified: the available information is not sufficient to unambiguously perform the task.

Imagine that a robot is going to grasp an object and manipulate it --- understanding the uncertainty over the size can have a profound impact on what the robot decides to do next.
For example, if it uses the predicted bounding box to avoid collisions during motion planning, then we may want to be conservative and err on the larger side.
However, if it is trying to pack the items into a shipment, then having accurate dimensions may also be important.

Consider the scene depicted in Figure \ref{fig:stacked_boxes}, which we observed in a real-world robotics application.
From the image of the object in a), it is fairly easy to gauge the width and length of the indicated object, but how tall is it? 
The object could be as deep as the bin, or it could be a stack of two identical objects, or even a thin object -- but from the available information, it is impossible to say for sure.
Formulating bounding box prediction as a regression problem results in a model that can only make a ``pointwise" prediction -- even in the face of ambiguity, we will only get a single predicted bounding box, shown in b).

A sufficiently expressive bounding-box model should be able to output the entire range of plausible bounding box hypotheses and make different predictions for different confidence requirements. 
A 0.5-confidence box d) must contain the object 50\% of the time while a 0.8-confidence box e) will expand in the direction of uncertainty to contain the object 80\% of the time. 
Moreover, such a model could leverage additional information, such as known dimensions of an object, to make even more accurate predictions, as shown in f).

Setting aside partial observability, the prediction space has complexities that require care in the design of a bounding-box estimator.
Making accurate predictions requires the estimator to reason about rotations, which has been observed to be notoriously difficult for neural networks to predict and model uncertainty over~\cite{rotation_cont_Zhou2019OnTC,Gilitschenski2020Deep,peretroukhin_so3_2020}. 
Many existing methods sidestep this problem by constraining their predictions to allow rotation about a single axis or no rotations at all.
This can be sufficient for some applications but has shortcomings for the general case.

A common thread that links these challenges together is the necessity to reason about uncertainty.
This has been largely underexplored in existing work, but we hypothesize that it is critical to improving 3D bounding box estimators and expanding their usability in applications of interest.
We propose to tackle this problem by predicting a more expressive probability distribution that explicitly accounts for the relationships between different box parameters.
Using a technique that has proven effective in other domains, we propose to model 3D bounding boxes autoregressively: that is, to predict each box component sequentially, conditioned on the previous ones.
This allows us to model multimodal uncertainty due to incomplete information, make high confidence predictions in the face of uncertainty, and seamlessly relax the orientation constraints that are popular in existing methods. To summarize our contributions:
\begin{enumerate}
    \item We propose an autoregressive formulation to 3D bounding box prediction that can model complex, multimodal uncertainty. We show how this formulation can gracefully scale to predict complete 3D orientations, rather than the 0- or 1-D alternatives that are common in prior work.
    \item We propose a method to make high confidence predictions in ambiguous scenarios and estimate useful measures of uncertainty.
    \item We introduce a simulated dataset of robotics scenes that illustrates why capturing uncertainty is important for 3D bounding box prediction, as well as the benefits and challenges of predicting full 3D rotations.
    \item We show that our formulation applies to both traditional 3D bounding box estimation and 3D object detection, achieving competitive results on popular indoor and autonomous driving datasets in addition to our dataset.
\end{enumerate}

\section{Related Work}

\textbf{3D Bounding-box Estimation: } 
Early work on 3D bounding box prediction \cite{mousavian20173d,qi2018frustum} assumes that object detection or segmentation has already been performed, and the bounding box predictor solely needs to identify a single 3D bounding box within a filtered point cloud.
In this paper, we refer to this task as \textit{3D bounding-box estimation}.
Much of this work focused on developing architectures to easily consume point cloud data, which often can be sparse and/or unstructured when obtained from real-world data. 

\textbf{3D Object Detection: } 
Recently, a number of methods~\cite{shi20193d,rukhovich2021fcaf3d,liu2021group,misra2021end,votenet,pvrcnn,qi2020imvotenet} have explored how to jointly perform object detection and 3D bounding box estimation, rather than treating them as two explicit steps.
This task is known as \textit{3D object detection} and is quickly gaining popularity over the decoupled detection and estimation tasks.
The main focus is on how to take the network architectures that have proven successful at the estimation task (which have strong inductive biases for operating on point clouds), and combine them with the architectures commonly used for the 2D object detection method (which are usually based on region proposals).

\textbf{Uncertainty Modeling in Object Detection: }
Uncertainty modeling has been studied in the context of 2D and 3D Object Detection~\cite{meyerlasernet,genfocalloss,Zhong2020UncertaintyAwareVB,Meyer2020LearningAU,probobject}. In many cases, these methods will use independent distributions, such as Gaussian or Laplace, to model uncertainty over box parameters such as corners, dimensions, and centers~\cite{he2019bounding,meyerlasernet,Choi2019GaussianYA}. While these distributions may capture some uncertainty for simple box parameterizations, they don't capture correlations across parameters and have yet to be proven on full 3D rotations.

\textbf{Autoregressive Models: } 
Deep autoregressive models are frequently employed across a variety of domains.
In deep learning, they first gained popularity for generative modeling of images \cite{van2016pixel,oord2016wavenet,vaswani2017attention}, since they can model long-range dependencies to ensure that pixels later in the autoregressive ordering are sampled consistently with the ones sampled earlier.
In addition to being applied to other high-dimensional data such as audio \cite{oord2016wavenet}, they have also been shown to offer precise predictions even for much lower-dimensional data, such as robot joint angles or motor torques \cite{metz2017discrete}.

\section{Autoregressive 3D Bounding Box Prediction}
\label{sec:ar_boxes}

3D bounding box estimation is typically formulated as a regression problem over the dimensions $d = (d_x, d_y, d_z)$, center $c = (c_x, c_y, c_z)$, and rotation $R = (\psi, \theta, \phi)$ of a bounding box, given some perceptual features $h$ computed from the scene, e.g. from an image and point cloud.
Prior work has explored various parametrizations and loss functions, but a notable salient feature to observe is that they all predict a \textit{pointwise} estimate of the bounding box: the model simply outputs all of the box parameters at once.
In 3D object detection, such regression is typically applied to every box within a set of candidates (or \textit{anchors}), and fits into a larger cascade that includes classifying which anchors are relevant and filtering out unnecessary or duplicate anchors.
In practice, this formulation can be greatly limiting, especially in the face of partial observability or symmetry.

\subsection{Autoregressive Modeling}

We propose to tackle this problem by autoregressively modeling the components of a 3D bounding box.
That is, for some ordering of the components (e.g. dimensions $\rightarrow$ center $\rightarrow$  orientation, or any permutation thereof), such a predictor will sequentially predict each component conditioned on the previous ones.
In theory, the particular autoregressive ordering should not matter; empirically, we find that dimensions $\rightarrow$ center $\rightarrow$ orientation was effective, so we use this ordering for our model.
Having dimension as first in the autoregressive ordering also enables us to condition on dimensions when they are known which can be effective at improving the prediction accuracy.

We discretize the box parameters rather than predicting continuous values, which is a well-known technique that allows the model to easily express multimodal distributions~\cite{van2016pixel}.
For rotations, we chose Euler angles since each dimension has a fixed range and does not to be normalized.
To make discrete dimension and center predictions, we normalize those parameters so that they can fit within a fixed set of discrete bins. 
We normalize dimensions by some scale $s$ so that most values of $d/s$ are within the range $[0, 1]$, and offset the centers by $c_0$ so that most normalized centers $(c-c_0)/s$ are within the range $[-1, 1]$. We use 512 bins for each dimension and adjust the bin range to achieve on average $\ge 0.99$ IOU with the quantized box and $<0.1\%$ overflow or underflow due to quantization.

From RGB-D inputs we extract a fixed-dimensional feature vector $h$ for each object. For each parameter $b = (d_x, d_y, d_z, c_x, c_y, c_z, \psi, \theta, \phi)$ in the autoregressive ordering, we model $p(b_i | b_1, \dots, b_{i-1}, h)$ using a MLP with 2-3 hidden layers. This autoregressive model is then trained using maximum likelihood:
\begin{align}
\log p(b|h) &= \sum_{i=1}^{9} \log p(b_i | b_1, \dots, b_{i-1}, h)
\end{align}

\begin{figure}[t]
    \centering
    \includegraphics[width=\textwidth]{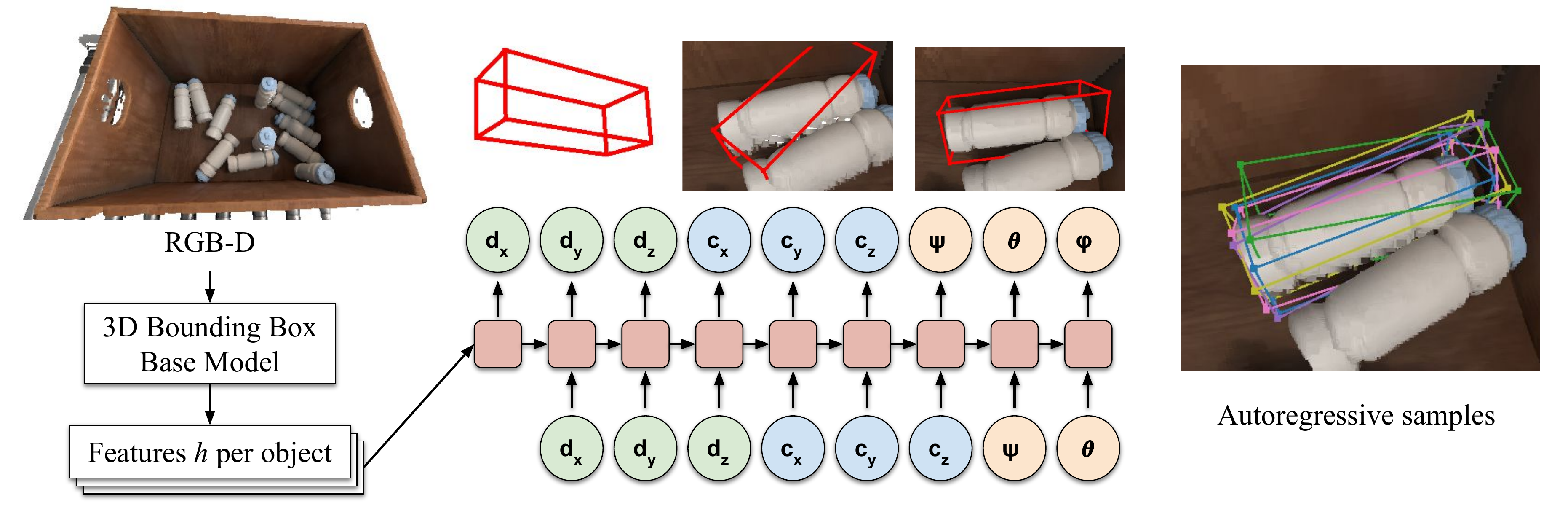}
    \caption{
    We compute per-object features $h$ using a base model from RGB-D input. Then, we autoregressively sample dimensions, center, and rotations, each step conditioned on the previous one. We can express uncertainty through samples, such as the rotational symmetry of the bottle, whereas pointwise models could only make a single prediction.
    }
    \label{fig:model_arch}
\end{figure}

\subsection{Model Architectures}
\label{sec:model_arch}

Our autoregressive prediction scheme can be applied to any type of 3D bounding box predictor.
In this section, we discuss how it might be applied in two different contexts: 3D object detection and 3D bounding box estimation.

\subsubsection{Autoregressive 3D Object Detection.}
\label{sec:model_arch_detection}
FCAF3D \cite{rukhovich2021fcaf3d} is a state-of-the-art 3D object detection method that was heavily engineered to exploit sparse and unstructured point clouds.
Given a colored point cloud, it applies a specialized feature extractor consisting of sparse 3D convolutions, and then proposes 3D bounding boxes following a popular single-stage detector, FCOS \cite{tian2019fcos}.

\begin{figure}[t]
\centering

\includegraphics[width=1.0\linewidth]{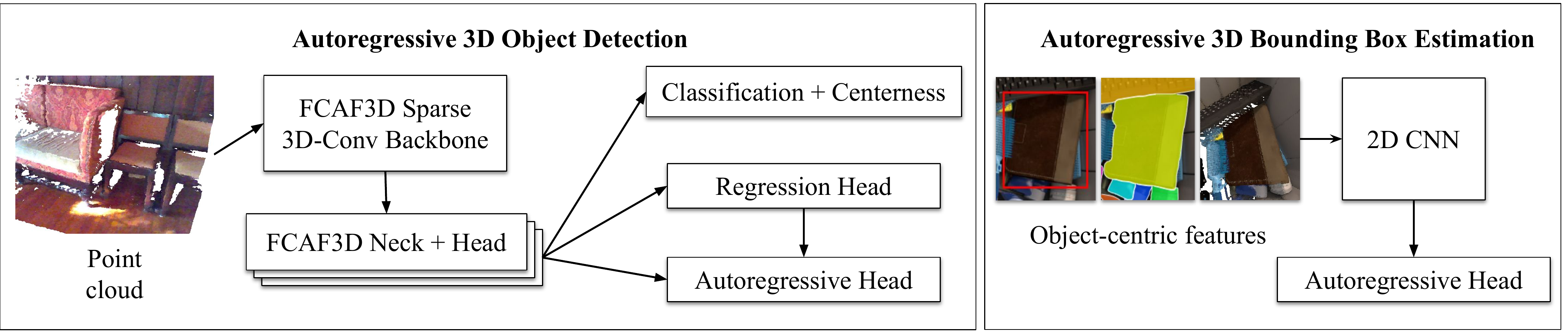}
\caption{For indoor 3D Object Detection, we use FCAF3D as a base model with an autoregressive head for bounding box prediction. For 3D Bounding Box Estimation we take object-centric features from a 2D object detector and pass them into a 2D CNN for autoregressive bounding box prediction.}
\label{fig:estimation-illustration}
\end{figure}

\textbf{Autoregressive FCAF3D:} We can make FCAF3D autoregressive by adding a head and training this head with maximum likelihood in addition to the FCAF3D loss $L_F(h, y)$ (Figure~\ref{fig:estimation-illustration}).
We found that the pointwise box prediction was useful to condition the autoregressive prediction and estimate the scaling normalization factor $s = \max\{d_x', d_y', d_z'\}$, where $d'$ is the pointwise dimension prediction of FCAF3D. 
Bounding box centers $c$ are normalized by the output locations $c_0$ of the sparse convolutions and scaled by the same $s$: $(c-c_0)/s$.
Since 3D object detection datasets have at most one degree of freedom for rotation, we predict only one $\theta$ parameter for box rotation.

To optimize the autoregressive prediction for higher IOU, we sample boxes $b\sim p(b|h)$ and maximize the IOUs of the samples with the ground truth box $y$. For this optimization, we use the conditional expectation $b'$ where $b'_i = \mathbb{E}[b_i | b_1, \dots, b_{i-1}, h]$ (since $b'$ is differentiable) to maximize $IOU(b', y)$. Altogether, we train autoregressive FCAF3D using the combined loss:
\begin{align}
    L(h, y) = L_F(h, y) - \log p (b | h) + \mathbb{E}_{b\sim p(b|h)}[1-IOU(b', y)]
\end{align}

\textbf{Autoregressive PV-RCNN:} Lidar-based object detection networks, such as PV-RCNN~\cite{pvrcnn}, typically have different architectures and inductive biases than indoor detection models. However, we show that our autoregressive box parameterization is agnostic to the underlying architecture by applying it to PV-RCNN. We propose Autoregressive PV-RCNN by extending the proposal refinement head to be autoregressive, modeling the residual $\Delta r^{\alpha}$ as discrete autoregressive $p(\Delta r^{\alpha} | h)$. Then, we add $-\log p(\Delta r^{\alpha} | h)$ to the total training loss.

\subsubsection{Autoregressive 3D Bounding Box Estimation.}
\label{sec:model_arch_estimation}

3D Bounding Box Estimation assumes that object detection has already been performed in 2D, and we simply need to predict a 3D bounding box for each detected object.
To highlight that our autoregressive prediction scheme can be applied to any bounding box predictor, we chose a model architecture that is substantially different from FCAF3D. 
For each detected object, we take an object-centric crop of the point cloud, normals, and object mask as input to a 2D-CNN, producing a fixed feature vector $h$ per object. 
This $h$ is used as features for our autoregressive parameterization $p(b|h)$.
See Appendix~\ref{appendix:model} for more details on the architecture.

To normalize the input and box parameters, we scale by the range of the first and third quartiles of each point cloud dimension $s = Q_3 - Q_1$, and recenter by the mean of the quartiles $c_0 = \frac{Q_1 + Q_3}{2}$. For full SO(3) rotations, we found there were many box parameters that could represent the same box; for example, a box with $d=(1,2,3)$ is equivalent to a box with $d'=(2,1,3)$ and a $90^{\circ}$ rotation. To account for this, we find all the box parameters $B = \{b^{(1)}, ..., b^{(m)} \}$ that represent the same box and supervise on all of them:
\begin{align}
    L(h, B) = -\frac{1}{|B|}\sum_{b^{(i)} \in B} \log(b^{(i)} | h)
\end{align}

\section{Applying Autoregressive 3D Bounding Box Models}

Given a trained autoregressive bounding-box model, how do we actually obtain predictions from it?
There can be a few different options, depending on how the downstream application plans to use the predictions.

\subsection{Beam Search}
In many applications, we want to simply obtain the most likely 3D bounding box given the input observation.
That is, we find the box $b^* = \arg\max_{b} p(b|h)$ which is most likely under the model.
Finding $b^*$ exactly can be computationally expensive, but we can approximate it using \textit{beam search}, a technique that has proven especially popular for autoregressive models in natural language applications~\cite{freitag-al-onaizan-2017-beam}.
Beam search allows us to estimate the mode of the distribution learned by the model and serves as an effective pointwise prediction.

\subsection{Quantile and Confidence Boxes} \label{sec:quantile-box}

In applications such as robotics and autonomous driving, 3D bounding boxes are often used to estimate object extents and avoid collisions. 
To that end, we often care that an object $o$ is fully contained in the estimated box $b$.
For a given confidence requirement $p$, we define a confidence box $b_p$ as a box that contains the true object $o$ with probability at least $p$: $\mathbb{P}(o \subseteq b_p ) \ge p$.
We'll show how to use an autoregressive bounding box model for confidence box predictions.

Suppose we draw multiple samples $K$ from our model. If a point $x \in \mathbb{R}^3$ is contained in many boxes, then it's likely that point is actually part of the object.
Conversely, a point that is only contained in a few sampled boxes is not likely to be part of the object.
We can formalize this intuition as the occupancy measure
\begin{align}
O(x) = \mathbb{P}(x \in b) = \mathbb{E}_{b \sim p(b|h)}[\mathbbm{1}\{ x \in b\}] 
\approx \frac{1}{K} \sum_{i=1}^{K}{ \mathbbm{1}\{ x \in b^{(i)}\} }
\end{align}
which can be approximated using samples $b^{(1)}, \dots, b^{(K)} \sim p(b|h)$ from our model.

To find regions that are very likely to be part of an object, consider the set of all points that have occupancy greater than $q$:
\begin{align}Q(q) &= \{ x : O(x) > q \}\end{align}
which we'll refer to as the \textit{occupancy quantile}. The minimum volume bounding box over the occupancy quantile is the \textit{quantile box}:
\begin{align}b_q &= \arg\min_{b: Q(q) \subseteq b} \text{vol}(b)\end{align}

\begin{figure}[t]
\centering
\includegraphics[width=\linewidth]{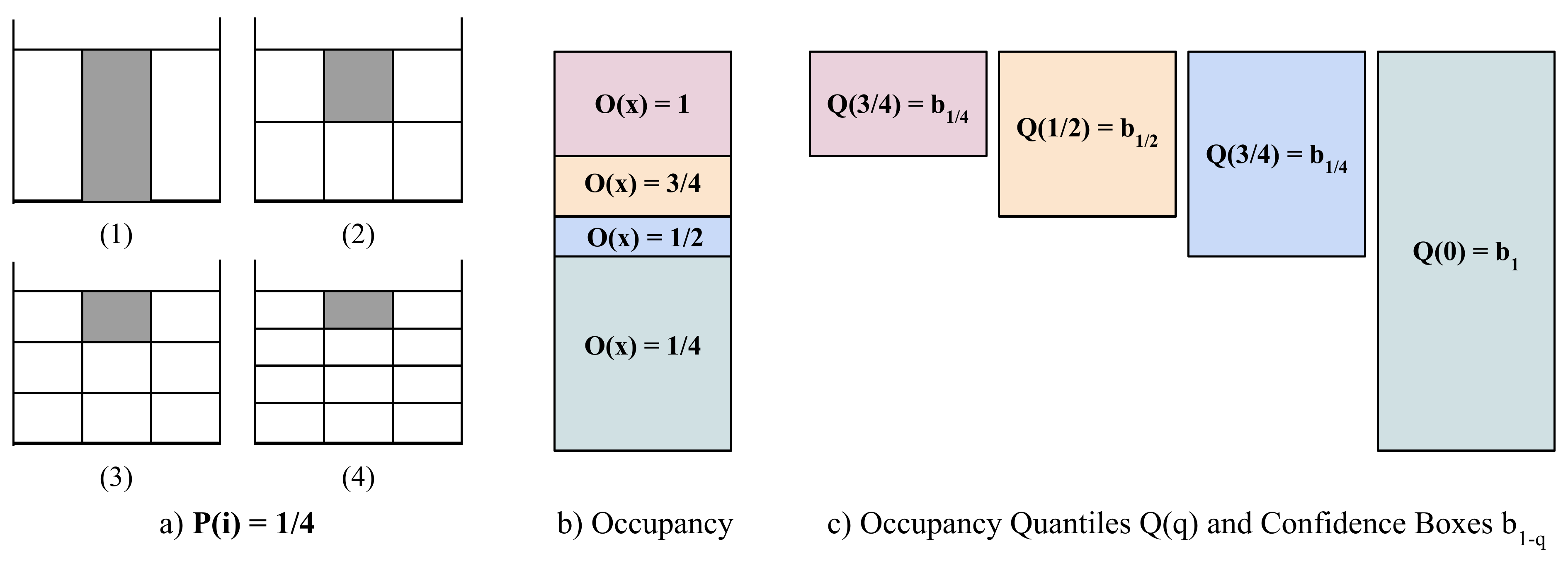}
\caption{Consider a scenario where we are estimating the bounding box of a tightly packed bin of stacked boxes. a) There is not enough visual information to estimate the object height, however, we know that the object could have heights $H/i$ for $i\in\{1,2,3,4\}$ with equal probability. b)  We compute the occupancy $O(x)$ for different regions. c) We visualize occupancy quantiles $Q(q)$ which correspond to confidence boxes $b_{1-q}$. Notice that as the confidence requirement increases, the size of the box increases to ensure we can contain the true object.}
\label{fig:quantile-illustration}
\end{figure}

Under some conditions, we can show that quantile boxes are confidence boxes.
\begin{theorem}
\label{thm:quantile}
A quantile box with quantile $q$ is a confidence box with confidence $p=1-q$ when $p(b|h)$ is an ordered object distribution. 
\end{theorem}
$p(b|h)$ is an ordered object distribution if for any two distinct boxes $b_i, b_j$ in the sample space of $p(b|h)$, one box must be contained within the other, $b_i \subset b_j$ or $b_j \subset b_i$. 
Empirically we find that quantile boxes are good approximations for confidence boxes even when $p(b|h)$ is not an ordered object distribution. 
See Figure~\ref{fig:quantile-illustration} for a visualization of occupancy and confidence boxes.

Quantile boxes provide an efficient way to make confidence box predictions with an autoregressive model.
We can use the autoregressive distribution to estimate occupancy using supervision from 3D box labels (without requiring meshes for direct occupancy supervision). 
Occupancy quantiles provide a fast approach for confidence box estimation on ordered object distributions and a good confidence box approximation for general object distributions. Appendix~\ref{appendix:quantile} has the full proof of Theorem~\ref{thm:quantile} and the details of our fast quantile box algorithm.

\subsection{Uncertainty Measure} \label{sec:uncertainty-measure}

Uncertainty estimation is an important application of bounding box estimation. When the 3D extent of an object is unknown or not fully observed, it can be valuable if a model can also indicate that its predictions are uncertain. For instance, a robot may choose to manipulate that uncertain object more slowly to avoid collisions, or an autonomous vehicle may be more cautious around a moving object of unknown size. 

A pointwise predictor can accomplish this by predicting both a mean $\mu$ and variance $\sigma^2$ for each box parameter, maximizing a $\mathcal{N}(\mu, \sigma^2)$ likelihood \cite{he2019bounding}. 
However, the spread of the distribution is measured independently for each box parameter which doesn't measure the spread of the overall box distribution well.

With an autoregressive box parameterization, we can measure uncertainty in the space of boxes using quantile boxes. Let $b_\alpha$ and $b_\beta$ be two quantile boxes with different quantiles. If we consider these boxes as confidence boxes, we can interpret $(b_\alpha, b_\beta)$ as a confidence interval or the spread of the box distribution. With this intuition, we can measure uncertainty using the IOU of different quantile boxes $U_{\alpha,\beta} = 1 - IOU(b_\alpha, b_\beta)$. This $U_{\alpha,\beta}$ effectively measures the span of the distribution in units of relative volume.

\subsection{Dimension Conditioning}

For some robotics applications, such as object manipulation in industrial settings, we are often presented with Stock-Keeping Unit, or SKU, information beforehand. In these scenarios, the dimensions of each SKU are provided, and the prediction task essentially boils down to correctly assigning the dimensions to a detected object instance, and predicting the pose of the 3D bounding box.

The autoregressive nature of our model allows for conveniently conditioning on the dimensions of each bounding box. However, we don't know which object in the scene corresponds to which SKU dimensions. How can we leverage dimension information from multiple SKUs without object-SKU correspondence? Our autoregressive model provides an elegant solution using conditioning and likelihood evaluation.

Given $\{d^{(1)}, ..., d^{(k)}\}$ known SKU dimensions, we can make a bounding box prediction using this information by maximizing:
\begin{align}
    b^* = \arg\max_b \{\max_{d^{(1)}, ..., d^{(k)}} p(b | d^{(i)}, h)\}
\end{align}
We can find the optimal $b^*$ by using beam search conditioned on each of the $d_i$ and returning the box with the highest likelihood.
Figure~\ref{fig:stacked_boxes} shows an illustrative example of how dimension conditioning can be used to greatly increase the fidelity of the predicted 3D bounding boxes.

\section{Experiments}
\label{sec:experiments}

We designed our experiments to answer the following questions:
\begin{itemize}
    \item[1.] How does an autoregressive bounding box predictor perform compared to a pointwise predictor, across a variety of domains and model architectures? 
    \item[2.] How meaningful are the uncertainty estimates from an autoregressive model? Are quantile boxes confidence boxes for general object distributions?
\end{itemize}

\subsection{Datasets}
To demonstrate the flexibility of our method, we conducted experiments on a diverse set of indoor, outdoor, and industrial datasets:

\textbf{SUN-RGBD} \cite{song2015sun} is a real-world dataset containing monocular images and point clouds captured from a stereo depth camera. 
It features a large variety of indoor scenes and is one of the most popular benchmarks in 3D object detection.
The box labels only include one rotational degree of freedom $\theta$. 

\textbf{Scannet} \cite{dai2017scannet} is a dataset of indoor 3D reconstructions. 
There are 18 classes and box labels are axis-aligned (no rotation). We train on 1201 scenes and evaluate on 312 validation scenes. 

\textbf{KITTI} \cite{kitti} is a widely popular 3D detection dataset for autonomous driving. Objects in KITTI have one degree of rotational freedom $\theta$, and we report evaluation results on the validation split. 

\textbf{COB-3D}. \textit{Common Objects in Bins 3D} is a simulated dataset rendered by \href{https://www.theorystudios.com/}{Theory Studios} to explore a qualitatively different set of challenges than the ones exhibited in popular datasets in the literature. 
We are releasing nearly 7000 scenes that aim to emulate industrial order-picking environments with each scene consisting of a bin containing a variety of items. 
There are two main themes we chose to highlight:
first, the objects are in a greater range of orientations than any other 3D-bounding-box dataset.
In particular, a model that performs well must reason about complete 3D rotations, whereas the state-of-the-art methods on SUN-RGBD only need to predict one rotational degree of freedom.
Secondly, it exhibits many types of ambiguity including rotation symmetry, occlusion reasoning in cluttered scenes, and tightly-pack bins with unobserved dimensions.
See Appendix~\ref{appendix:dataset} for full details on this dataset including visual examples.

\subsection{Evaluation}
To evaluate 3D-bounding-box predictions, \textit{intersection-over-union}, or IoU, is commonly used to compare the similarity between two boxes.
3D object detection uses \text{mean average precision}, or mAP, to measure how well a detector trades off precision and recall.
IoU is used to determine whether a prediction is close enough to a ground-truth box to constitute a true positive.
For 3D bounding-box estimation, detection has already happened, so we simply measure the mean IoU between the prediction and ground-truth, averaged across objects.

Unlike 2D detection, many applications that use 3D bounding boxes especially care about underestimation more than overestimation: if the predicted bounding box is too large, that is generally a less costly error than if it is too small. In the latter case, there are parts of the object that are outside the bounding box, which may result in collisions in robotics or autonomous driving setting.

To help quantify this error asymmetry, we consider a new similarity functions, the \textit{intersection-over-ground-truth} (IoG).
IoG measure what fraction of the ground truth box is contained within the predicted box; when IoG is 1, the ground truth box is fully contained in the predicted box.
With IoG and IoU, we have a more complete understanding of the types of errors that a bounding-box predictor is making. 
For the detection task, we compute mAP separately using IoU and IoG, and for the estimation task, we compute the mean IoG along with the mean IoU.

\begin{table}[t]
\begin{center}
\caption{3D Object Detection results on SUN-RGBD, Scannet, and KITTI}
\label{table:sunrgbd-metrics}
\begin{tabular}{|c|l|ccc|ccc|}\hline
 & &\multicolumn{3}{c|}{IoU}&\multicolumn{3}{c|}{IoG}\\
 Dataset & Method &$AP_{0.25}$&$AP_{0.50}$&$AP_{all}$&$AP_{0.25}$&$AP_{0.50}$&$AP_{all}$\\\hline
\parbox[t]{2mm}{\multirow{9}{*}{\rotatebox[origin=c]{90}{SUN-RGBD}}} &FCAF3D &  63.8& 48.2& 37.42& 64.72& 59.82& 48.75\\
&3DETR & 59.52 & 32.17 & 31.13 & 63.00 & 53.33 & 44.08\\
&VoteNet & 60.71& 38.98& 30.25& 62.81& 54.58& 43.62\\
&ImVoteNet & \textbf{64.24}& 39.38& 31.12& \textbf{67.00}& 57.41& 45.78\\
&Beam Search & \cellcolor{red!25}62.94& \cellcolor{red!25}47.03& \cellcolor{green!25}38.15& \cellcolor{red!25}64.75& \cellcolor{red!25}58.50& \cellcolor{red!25}47.17\\
&Quantile 0.1 & \cellcolor{red!25}61.21& \cellcolor{red!25}30.94& \cellcolor{red!25}31.06& \cellcolor{red!25}65.89& \cellcolor{green!25}\textbf{64.34}& \cellcolor{green!25}\textbf{60.08}\\
&Quantile 0.4 & \cellcolor{red!25}63.46& \cellcolor{green!25}48.41& \cellcolor{green!25}38.43& \cellcolor{red!25}65.34& \cellcolor{green!25}61.68& \cellcolor{green!25}51.76\\
&Quantile 0.45 & \cellcolor{red!25}63.47& \cellcolor{green!25}\textbf{48.64}& \cellcolor{green!25}\textbf{38.55}& \cellcolor{red!25}65.19& \cellcolor{green!25}61.03& \cellcolor{green!25}50.36\\
&Quantile 0.5 & \cellcolor{red!25}63.30& \cellcolor{red!25}47.70& \cellcolor{green!25}38.50& \cellcolor{red!25}64.99& \cellcolor{green!25}59.83& \cellcolor{red!25}48.44\\
    \hline
    
\parbox[t]{2mm}{\multirow{6}{*}{\rotatebox[origin=c]{90}{Scannet}}} & FCAF3D & 68.53& \textbf{53.87}& 43.32& 72.05& 67.63& 60.66\\
& 3DETR & 64.09& 47.16& 39.57& 68.62 & 59.17 & 49.82\\

& Beam Search & \cellcolor{green!25}\textbf{69.06}& \cellcolor{red!25}53.67& \cellcolor{green!25}\textbf{43.85}& \cellcolor{red!25}71.46& \cellcolor{red!25}66.10& \cellcolor{red!25}59.13\\
& Quantile 0.1 & \cellcolor{red!25}67.10& \cellcolor{red!25}43.13& \cellcolor{red!25}34.17& \cellcolor{green!25}72.23& \cellcolor{green!25}\textbf{70.01}& \cellcolor{green!25}\textbf{66.73}\\
& Quantile 0.2 & \cellcolor{red!25}68.03& \cellcolor{red!25}48.68& \cellcolor{red!25}38.27& \cellcolor{green!25}\textbf{72.30}& \cellcolor{green!25}69.68& \cellcolor{green!25}65.43\\
& Quantile 0.4 & \cellcolor{green!25}68.73& \cellcolor{red!25}52.98& \cellcolor{red!25}42.76& \cellcolor{green!25}72.08& \cellcolor{green!25}67.74& \cellcolor{green!25}61.98\\
\hline

\parbox[t]{2mm}{\multirow{7}{*}{\rotatebox[origin=c]{90}{KITTI}}} &&\multicolumn{3}{c|}{AP IoU Hard Split}&\multicolumn{3}{c|}{AP IoG Hard Split}\\
 & Method & Car & Ped. & Cycl.  & Car & Ped. & Cycl.\\
\cline{2-8}
& PVRCNN & \textbf{82.37} & 53.12 & 68.69 & 91.86 & 67.08 & 73.14\\
& Beam Search & \cellcolor{green!25}\textbf{82.37}& \cellcolor{red!25}52.28& \cellcolor{green!25}\textbf{69.13}& \cellcolor{red!25}91.84& \cellcolor{red!25}66.96& \cellcolor{green!25}73.40\\
& Quantile 0.1 & \cellcolor{red!25}59.75& \cellcolor{red!25}39.26& \cellcolor{red!25}58.38& \cellcolor{green!25}\textbf{96.02}& \cellcolor{green!25}\textbf{71.85}& \cellcolor{green!25}\textbf{76.09}\\
& Quantile 0.4 & \cellcolor{red!25}81.98& \cellcolor{green!25}\textbf{54.15}& \cellcolor{red!25}68.45& \cellcolor{green!25}93.98& \cellcolor{green!25}70.63& \cellcolor{green!25}74.08\\
& Quantile 0.5 & \cellcolor{red!25}82.32& \cellcolor{green!25}53.78& \cellcolor{green!25}69.03& \cellcolor{red!25}91.84& \cellcolor{green!25}68.14& \cellcolor{green!25}73.52\\
    \hline

\end{tabular}
\end{center}
\end{table}

\subsection{3D Object Detection}

To evaluate the autoregressive box parameterization for 3D Object Detection, we evaluate Autoregressive FCAF3D and Autoregressive PV-RCNN introduced in Section~\ref{sec:model_arch_detection}. Table~\ref{table:sunrgbd-metrics} shows the comparison between autoregressive models and baselines on SUN-RGBD, Scannet, and KITTI. We find that beam search generally matches the baseline performance, if not exceeding performance on IoU $AP_{all}$.

As for quantile boxes, we find that lower quantiles result in higher IoG mAP which suggests that the predicted boxes are more likely to contain the ground truth box. This is consistent with our claim from Theorem~\ref{thm:quantile} since lower quantiles correspond to higher confidence boxes and must contain the true object with higher probability. We find that quantile boxes 0.4-0.5 strike the best balance between IoU and IoG, achieving better mAP than baselines in most cases. This flexible quantile parameter enables applications to trade off bounding box accuracy as measured by IoU with containment probability as measured by IoG. For instance, an autonomous vehicle may use a lower quantile to mitigate the risk of collisions at the cost of some bounding box accuracy. 

\subsection{3D Bounding Box Estimation}

We evaluate the bounding box estimation on COB-3D using the model architecture described in Section~\ref{sec:model_arch_estimation}. To compare the effectiveness of our autoregressive parameterization, we train the same model architecture with different box parameterizations and losses. All models receive the same 2D detection results and features as input and must make 3D bounding box predictions for each detected object. We consider 4 baseline parameterizations for this task inspired by various works in the literature:

\textbf{L1 Regression:} 
In this parameterization, the model outputs 9 real values for each of the 9 box parameters:
$b = (d_x, d_y, d_z, c_x, c_y, c_z, \psi, \theta, \phi)$.
The model predicts dimensions and centers in coordinates normalized around the object's point cloud. 
This model is trained using a L1 loss over the normalized box parameters $L(b, g) = ||b-g||_1$, where $g$ is the ground truth box~\cite{misra2021end}.

\textbf{Gaussian:}
For this baseline, the model outputs 18 real values for the mean, $\mu$, and log-variance, $\log\sigma^2$, of 9 Gaussian distributions $\mathcal{N}(\mu, \sigma)$ over the box parameters $b$~\cite{he2019bounding,meyerlasernet}.
Predicting the variance enables the model to output uncertainty over different box parameters, independently of each other. 
We train this model using maximum likelihood: $L(\mu, \log\sigma^2, g) = -\sum_i \log \mathcal{N}(g_i; \mu_i, \sigma_i)$.

\textbf{Discrete:}
In some prior works, box parameters are predicted as discrete bins but not in an autoregressive manner~\cite{conf/iccv/QiLHG19}. 
To evaluate this parameterization and ablate the necessity of autoregressive predictions, we predict each box parameter \textit{independently} as discrete bins:
$\log p(b|h) = \sum_{i=1}^9 \log p(b_i | h)$

\textbf{4-Point:}
This baseline outputs 12 real values for four 3D corner points $(p_0, p_1, p_2, p_3) \in \mathbb{R}^3$, constituting a 3D bounding box~\cite{meyerlasernet,Meyer2020LearningAU}. We ensure that the 3D bounding box is orthogonal by applying the Gram-Schmidt process on the basis vectors $(p_1 - p_0, p_2 - p_0, p_3 - p_0)$. We use an L1-loss on the difference between the predicted points and the points of the ground truth 3D bounding box. Since there are many permutations of valid 4-point corners of a bounding box, we supervise on the permutation that induces the minimum loss.

\begin{table}[t]
\begin{center}
\caption{
    Results of the proposed method \& baselines on our dataset. We also show results for conditioning our method on ground truth dimensions}
\label{table:theory-metrics}
  \begin{tabular}{|l|c|c|c|c|c|c|}
    \hline
  &IoU & IoG & F1 &$err_{dim}[m]$ & $err_{quat}[rad]$ &  $err_{center}[m]$
  \\
    \hline
L1 Regression & 0.4219 & 0.6113 & 0.4992 &  0.0436 & 0.4667 & 0.0138\\
Discrete & 0.5232 & 0.6282 & 0.5709 &  0.0339 & 0.2926 & \textbf{0.0105}\\
Gaussian & 0.3169 & 0.5304 & 0.3967 &  0.0450 & 0.5154 & 0.0119\\
4-Point & 0.5688 & 0.7113 & 0.6321 & 0.0332 & 0.1999 & 0.0132\\
    \hline
Beam Search & \cellcolor{green!25}\textbf{0.6296} & \cellcolor{green!25}0.7877 & \cellcolor{green!25}0.6999 & \cellcolor{green!25}\textbf{0.0287} & \cellcolor{green!25}\textbf{0.1598} & \cellcolor{red!25}0.0109\\
Quantile 0.1 & \cellcolor{red!25}0.3821 & \cellcolor{green!25}\textbf{0.9723} &  \cellcolor{red!25}0.5486 & \cellcolor{red!25}0.0986 &\cellcolor{green!25}0.1762 & \cellcolor{red!25}0.0123\\
Quantile 0.4 & \cellcolor{green!25}0.5949 & \cellcolor{green!25}0.8871 & \cellcolor{green!25}0.7122 & \cellcolor{red!25}0.0377 & 0.\cellcolor{green!25}1640 & \cellcolor{red!25}0.0110\\
Quantile 0.5  & \cellcolor{green!25}0.6275 & \cellcolor{green!25}0.8295 & \cellcolor{green!25}\textbf{0.7126} &  \cellcolor{green!25}0.0318 & \cellcolor{green!25}0.1657 &\cellcolor{red!25}0.0110\\
    \hline
Conditioning & 0.6709 & 0.7899 & 0.7215 & 0.0086 & 0.1674 & 0.0096\\
    \hline
\end{tabular}
\end{center}
\end{table}

\subsubsection{Metrics.}

To make reasoning about the trade-off between IoG and IoU more quantifiable, we report the F1-score equivalent for this use case, i.e., $\text{F1}_{score} = \frac{2(IoU * IoG)}{IoU + IoG}$. We further report metrics on the dimension \& pose errors, which are computed as follows:
\begin{itemize}
\item[$\circ$] $err_{dim} = \text{sum}(|\mathbf{d} - \mathbf{d}_{gt}|)$, where we compute the error across all possible permutations and then choose the one with the smallest error.
\item[$\circ$] $err_{quat} = 2\,\text{arccos}(|\langle\mathbf{q}, \mathbf{q}_{gt}\rangle|)$, where \textbf{q} represents the rotational part of the pose as a quaternion. We compute the error across all possible symmetries and choose the one with the smallest error.
\item[$\circ$] $err_{center} = ||\mathbf{c} - \mathbf{c}_{gt}||_2$, where \textbf{c} is the 3D-center of the bounding box.
\end{itemize}

\subsubsection{Results.}
Table~\ref{table:theory-metrics} shows how our autoregressive methods compare to the baseline parameterizations. We find that \textit{Beam Search} achieves the best IoU, dimension \& rotation error.
As for the \textit{Quantile} methods, we find that lower quantiles achieve higher IoG while sacrificing IoU and dimension error.
\textit{Quantile 0.5} offers the best tradeoff in terms of overall performance, achieving higher IoG with similar IoU and dimension error compared to \textit{Beam Search}. 
Baseline models that predict box parameters directly generally performed worse since those models cannot properly capture multimodal correlations across the box parameters. The \textit{Discrete} baseline performs the best in terms of center error, but we can see that the best autoregressive methods are only a few millimeters worse.
For bounding box predictions with full rotations in SO(3), we find that an autoregressive bounding box parameterization can effectively model rotation uncertainty, achieving the lowest rotation error. We can also see that conditioning the model on known dimensions of the items in the scene increases performance in all relevant metrics (besides IoG), most notably in IoU \& dimension error. Note that the dimension error is non-zero because the model is given the dimensions as an unordered set, and still needs to predict the association of each dimension tuple to the corresponding item in the scene.\\

\subsection{Quantile and Confidence Boxes}

In Section~\ref{sec:quantile-box} we introduced quantile boxes as a fast approximation for confidence boxes. We showed that when $p(b|h)$ is an ordered object distribution, a quantile box with quantile $q$ is equivalent to a confidence box with $p=1-q$ and should contain the true object with probability $p$. 

\begin{figure}[t]
\begin{minipage}{\textwidth}
\begin{minipage}[b]{0.5\textwidth}
\centering
\includegraphics[width=\linewidth]{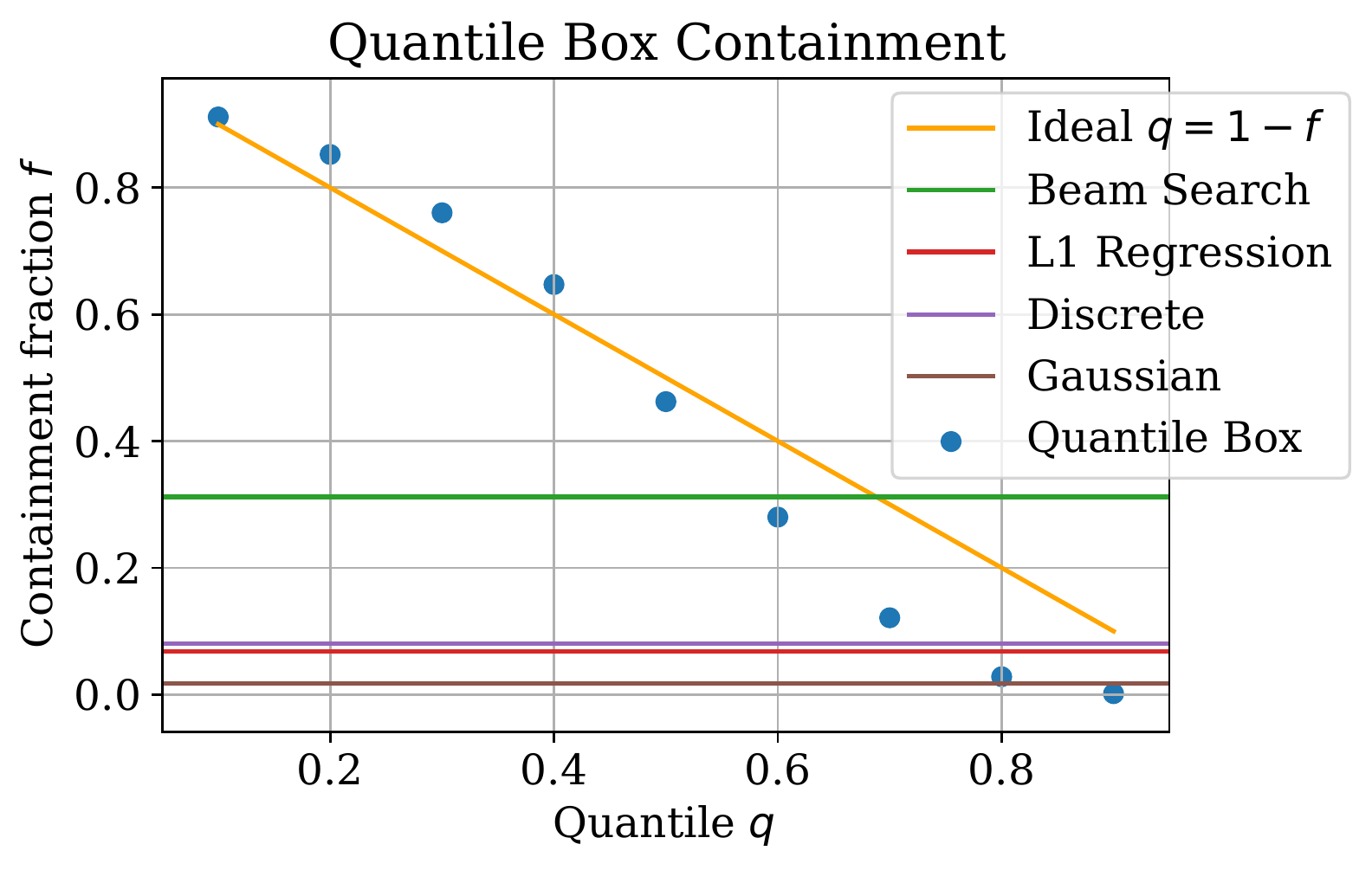}
\captionof{figure}{We compare fraction of predicted boxes that contain ground truth boxes $f$ with different quantiles $q$ and find that $q$-quantile boxes contain approximately $f\approx 1-q$ fraction of ground truth boxes.}
\label{fig:quantile-containment}
\end{minipage}
\hfill
\begin{minipage}[b]{0.46 \textwidth}
\centering
\begin{tabular}{|l|c|c|}
\hline
Method & ROC AUC & Spearman $r_s$\\
\hline
Gaussian & 0.731 & -0.530\\
Quantile 0.2 & 0.897 & \textbf{-0.865}\\
Quantile 0.5 & 0.878 & -0.789\\
Quantile 0.8 & \textbf{0.967} & -0.850\\
\hline
\end{tabular}
  \captionof{table}{We compare Quantile Uncertainty Measure $U_{0.2, 0.8}$ with Gaussian dimension variance $G$, and find that $U_{0.2, 0.8}$ a better predictor of ground-truth IoU compared to $G$ as measured by ROC AUC. $U_{0.2, 0.8}$ is also better correlated with ground-truth IoU compared to $G$ as measured by Spearman $r_s$} 
\label{table:uncertainty-measures}
\end{minipage}
\end{minipage}
\end{figure}

While it's hard to ensure  that real world objects follow an ordered distribution, we can empirically evaluate whether $q$ confidence boxes contain the ground truth object $1-q$ fraction of the time. To test our hypothesis, we predict quantile boxes with different $q$ and calculate the fraction of predictions $f$ with IoG $> 0.95$. In Figure~\ref{fig:quantile-containment}, we can see that $f \approx 1-q$ and follows a generally linear relationship. This suggests that even for general object distributions, quantile boxes can be an effective approximation for confidence boxes. 

\subsection{Uncertainty Measures}

In Section~\ref{sec:uncertainty-measure}, we introduced the uncertainty measure using quantile boxes $U_{\alpha, \beta}=1-IoU(b_{\alpha}, b_{\beta})$ as a measure of the span of the confidence box interval. 
To evaluate the effectiveness of this uncertainty measure, we calculate the ROC AUC of using $U_{0.2, 0.8}$ to predict when the IoU of the predicted box $b$ with the ground truth box $g$ is less than 0.25. 
We also measure the correlation between ground truth IoU and uncertainty using the Spearman's rank correlation $r_s$.
We compare $U_{0.2, 0.8}$ on different quantile boxes against Gaussian dimension variance $G  = \frac{\sigma_{d_x}\sigma_{d_y}\sigma_{d_z}}{\mu_{d_x}\mu_{d_y}\mu_{d_z}}$ on the Gaussian baseline.
Table~\ref{table:uncertainty-measures} shows that quantile uncertainty $U_{0.2, 0.8}$ can be a better uncertainty measure than $G$.

\section{Discussion}

We introduced an autoregressive formulation to 3D bounding prediction that greatly expands the ability of existing architectures to express uncertainty.
We showed that it can be applied to both the 3D object detection and 3D bounding-box estimation settings, and explored different ways to extract bounding box predictions from such autoregressive models.
In particular, we showed how the uncertainty expressed by these models can make high confidence predictions and meaningful uncertainty estimates.
We introduced a dataset that requires predicting bounding boxes with full 3D rotations, and showed that our model naturally handles this task as well.
While autoregressive models are just one class of distributionally expressive models, they are not the only option for more expressive bounding box modeling. 
We hope that future lines of work will continue to build upon the method, dataset, and benchmarks we introduced in this paper.

\clearpage
\bibliographystyle{splncs04}
\bibliography{egbib}

\appendix

\newpage

\section{Model Architecture and Training}\label{appendix:model}

\begin{figure}[t]
\centering
\includegraphics[width=\linewidth]{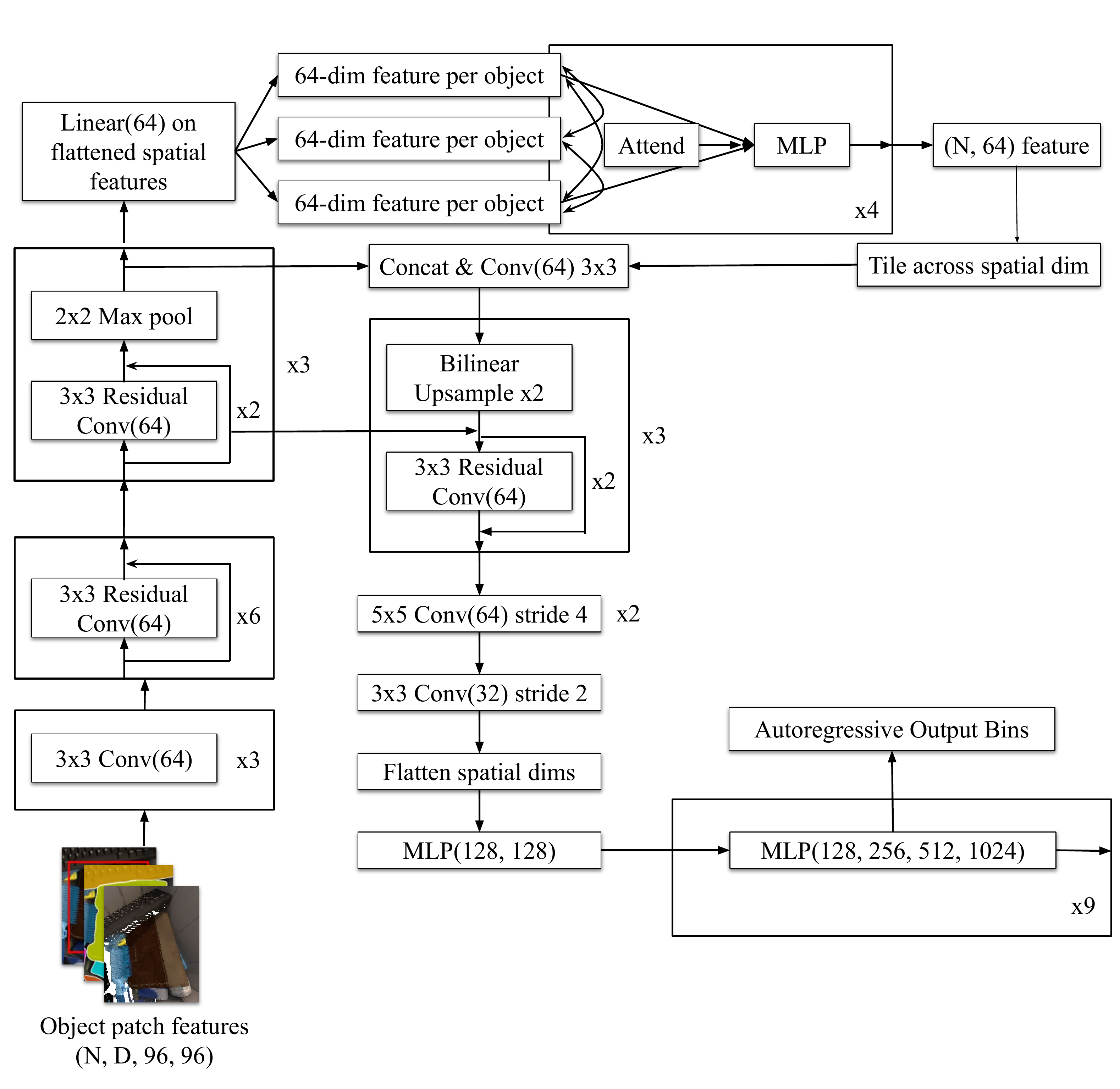}
\caption{Overview of Autoregressive Bounding Box Estimation architecture}
\label{fig:2d-arch}
\end{figure}

\subsection{Autoregressive 3D Bounding Box Estimation}

For bounding box estimation, our model operates on 2D detection patch outputs of size 96 x 96. We take the 2D bounding box from  object-detection to crop and resize the following features for each object: 3D point cloud, depth uncertainty score, normals, instance mask, amodal instance mask (which includes the occluded regions of the object). We normalize each point $p$ in the point cloud with the 0.25 ($Q_1$) and 0.75 ($Q_3$) quantiles per dimension using $\frac{p-c_0}{s}$ for $c_0=\frac{Q_1+Q_3}{2}$, $s=Q_3-Q_1$. We omitted RGB since we found it wasn't necessary for training and improved generalization.

We stack each 2D feature along the channel dimension and embed the features using a 2D Resnet U-Net. The features from the top of the U-Net are used in a series of self-attention modules across embeddings from all objects in a scene so that information can be shared across objects. The resulting features from self-attention are tiled across the spatial dimension before the downward pass of the U-Net. Finally, the features from the highest spatial resolution of the U-Net are passed into several strided-convs, flattened, and projected to a 128-dimension feature $h$ per object. Figure~\ref{fig:2d-arch} shows the overview of our model architecture.

For the autoregressive layers, we use 9 MLPs with hidden layers (128, 256, 512, 1024). 
For baselines, we keep the same architecture through $h$ and use different sized MLPs depending on the box parameterization.  
We train using Adam with learning rate 1e-5 with a batch size of 24 scenes per step with varying number of objects per scene.
We train for 10000 steps or until convergence.

\subsection{Autoregressive 3D Object Detection}

For Autoregressive FCAF3D, we add 7 autoregressive MLPs with hidden dimensions (128, 256, 512). All other parameters of FCAF3D are the same and we train the same hyperparameters as the released code for 30 epochs. For the baseline FCAF3D, we trained the author-released model for 30 epochs on 8 gpus. We found that the benchmarked numbers for $AP_{0.25}$ and $AP_{0.50}$ were slightly lower than the reported ones in the original paper, so in our table, we use the reported average $AP$ across trials from the original paper. $AP_{all}$ was calculated in a similar way as in MS-COCO by averaging $AP$ for iou thresholds over $0.05, 0.10, 0.15, ..., 0.95$.

\section{Quantile Box} \label{appendix:quantile}

\subsection{Proof of Quantile-Confidence Box}

\subsubsection{Proof Sketch:} Let $P(b)$ be a distribution over an ordered set of boxes where for any two distinct boxes $b_1, b_2$ in the sample space, one must be contained in the other, $b_1 \subset b_2$ or $b_2 \subset b_1$. We'll show that a quantile box $b_q$ is a confidence box with $p=1-q$ by 1) constructing a confidence box $b_p$ for any given $q$, 2) showing that any $x\in b_p$ must have $O(x) > q$, and 3) therefore $b_p \subseteq Q(q) \subseteq b_q$ so the quantile box is a confidence box.

\subsubsection{1) Confidence Box:} For any $p=1-q$, we'll show how to construct a confidence box $b_p$. Using the ordered object distribution property of $P(b)$, we can define ordering as containment $b_1 < b_2 \equiv b_1 \subset b_2$. This ordering defines an inverse cdf:
\begin{align}
    F^{-1}(p) = \inf\{ x: P(b \le x) \ge p \}
\end{align}
Let $b_p = F^{-1}(1-q)$ be the inverse cdf of $p$; by definition $b_p$ is a confidence box with confidence $p$ since $P(b \le b_p) = P(b \subseteq b_p) \ge p$

\subsubsection{2) Occupancy of $b_p$:} We'll show that any $x \in b_p$ satisfies $O(x) > 1-p$. First we'll prove that that $ P(b \ge b_p) > 1-p$. Let $b_0 = \inf\{ b : b < b_p \}$, the smallest box that is strictly contained in $b_p$. (If no such $b_0$ exists, then $b_p$ must be the smallest box in the distribution order such that $P(b \ge b_p)=1$ and $ P(b \ge b_p) > 1-p$ for $p\neq 0$)

Since $b_p$ is the inverse cdf of $p$, we know that $P(b \le b_0) < p$, otherwise $b_0$ would be the inverse cdf of $p$ (i.e. $b_0=b_p$ a contradiction). It follows that
\begin{align}
P(b \ge b_p) & = P(b > b_0)\\
 & = 1 - P(b \le b_0)\\
&> 1-p 
\end{align}

Now consider any point $x \in b_p$:
\begin{align}
    O(x) &= P(x\in b)\\
    &= \int_{b} \mathbbm{1}\{ x \in b\} p(b)db \\ 
    &\ge \int_{b \ge b_p } \mathbbm{1}\{ x \in b\} p(b)db \label{eq:noneg}\\ 
    &= \int_{b \ge b_p } p(b)db\label{eq:contain}\\ 
    &= P( b \ge b_p  )\\ 
    &> 1-p 
\end{align}
Where (\ref{eq:noneg}) follows from the nonegativity of $\mathbbm{1}\{ x \in b\} p(b)$. (\ref{eq:contain}) follows from $x\in b_p$, $b_p \subseteq b $ which implies $x \in b$. 

\subsubsection{3) Quantile-Confidence Box:}
Since any $x \in b_p$ satisfies $O(x) > 1-p$, it follows that $b_p \subseteq Q(1-p)$, where $Q(q) = \{ x : O(x) > q \}$ is the occupancy quantile with quantile $q$. The quantile box by construction must contain the occupancy quantile $Q(q) \subseteq b_q$, therefore we have $b_p \subseteq Q(1-p) \subseteq b_q$, and
\begin{align}
    P(b\subseteq b_q) &\ge P(b\subseteq b_p)\\
    &\ge p
\end{align}
So $b_q$ is a confidence box with confidence requirement $p$.

\subsection{Quantile Box Algorithm}

\begin{algorithm}[t]
\caption{Quantile Box Algorithm}
\label{algo:quantile}
\textbf{Given:} quantile $q$, box distribution $P(b|h)$, numbers of box samples $k$, number of point samples $m$\\
 Sample $b^{(1)}, ..., b^{(k)} \sim P(b|h)$ boxes\\
 For each $b^{(i)}$, sample $m$ random points within $b^{(i)}$, adding all points to a set $T$\\
 For all $x\in T$, estimate $O(x) = \frac{1}{k}\sum_i^k \mathbbm{1}\{x\in b^{(i)}\}$\\
 Construct the occupancy quantile $Q(q) = \{ x \in T : O(x) > q \}$\\
 \For{$b^{(1)}, ..., b^{(k)}$}{
    Let $R_i$ be the rotation of $b^{(i)}$\\
    Compute the volume of the $Q(q)$ bounding box under $R_i$, 
    $v_i = \prod_{a\in x,y,z} ( \max_{x\in Q(q)} (R_i^{-1} x)_a -  \min_{x\in Q(q)} (R_i^{-1} x)_a )$\\
  }
  Find the minimum volume box $i^* = \arg\min_i v_i$\\
  Let $s_a = \max_{x\in Q(q)} (R_{i^*}^{-1} x)_a$, $t_a = \min_{x\in Q(q)} (R_{i^*}^{-1} x)_a$\\
  Return box $b= (d, c, R_{i^*})$ with dimensions $d=(t_x-s_x, t_y-s_y, t_z-s_z)$ and center $c=R_{i^*}(s_x + d_x/2, s_y + d_y/2, s_z + d_z/2)$
 
\end{algorithm}

We propose a fast quantile box Algorithm~\ref{algo:quantile} that runs in polynomial time and is easily batchable on GPU. We use a finite sample of $k$ boxes to approximate the occupancy and a sample of $km$ points to approximate the occupancy quantile $Q(q)$. To find the minimum volume box, we assume that one of the sampled box rotations will be close to the optimal quantile box rotation. We take the sampled rotations and calculate the rotation-axis-aligned bounding box volume for the occupancy quantile. The minimum volume rotation is selected for the quantile box and corresponding dimension/center calculated accordingly.

Empirically we find that $k=64$, $m=4^3$ provides a good trade-off of variance and inference time. We can efficiently batch all operations on GPU, and find that quantile box inference for 15 objects takes no more than 10ms on a NVIDIA 1080TI.

\section{Dataset} \label{appendix:dataset}

\begin{figure}
\centering
\includegraphics[width=0.9\linewidth]{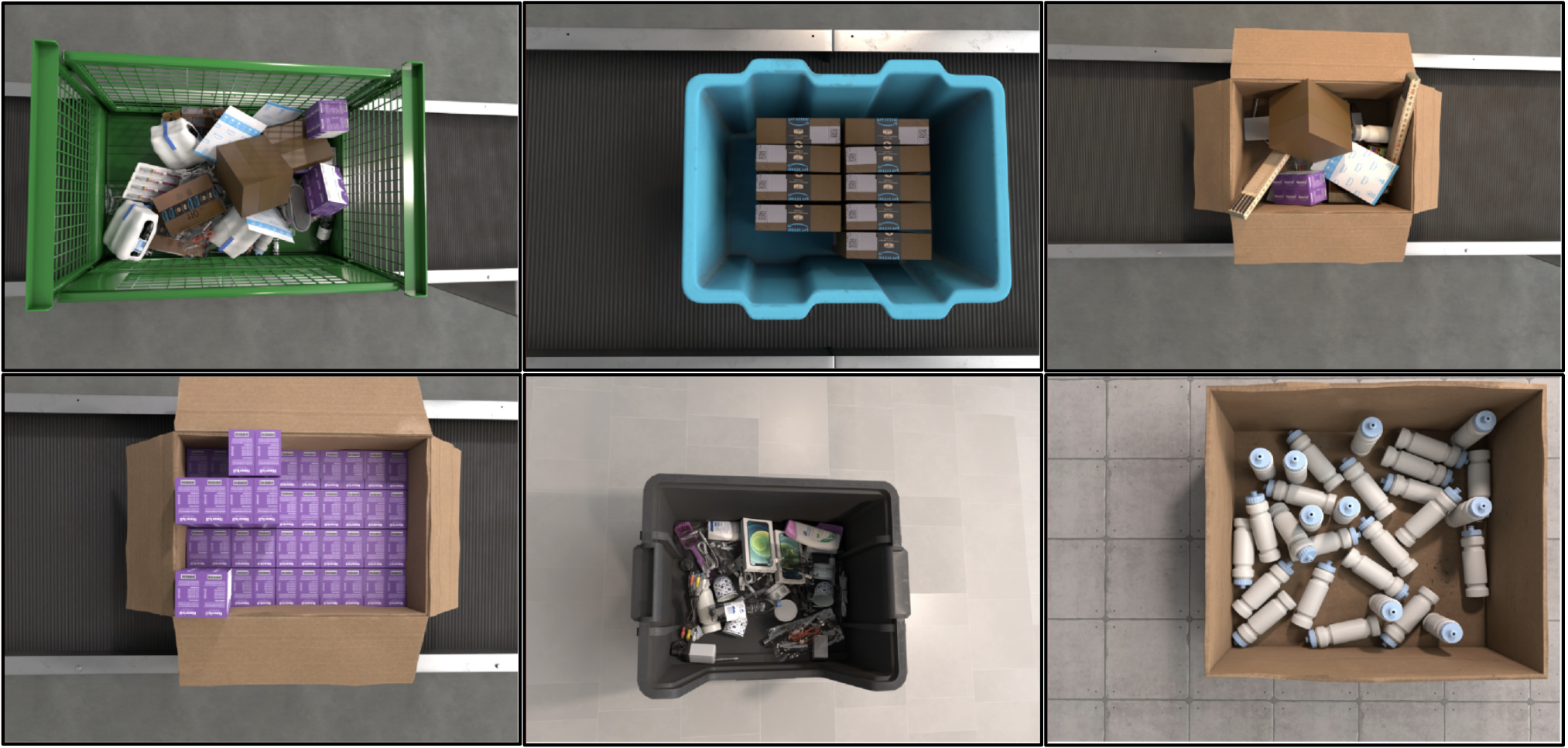}
\caption{Examples of scenes from our dataset}\label{fig:ts-examples}
\end{figure}

Our dataset consists of almost 7000 simulated scenes of common objects in bins. See Figure~\ref{fig:ts-examples} for examples. Each scene consists of the following data:
\begin{itemize}
    \item \textbf{RGB} image of shape (H, W, 3)
    \item \textbf{Depth} map of shape (H, W)
    \item \textbf{Intrinsic Matrix} of the camera (3, 3)
    \item \textbf{Normals Map} of shape (H, W, 3)
    \item \textbf{Instance Masks} of shape (N, H, W) where N is the number of objects
    \item \textbf{Amodal Instance masks} of shape (N, H, W) which includes the occluded regions of the object
    \item \textbf{3D Bounding Box} of each object (N, 9) as determined by dimensions, center, and rotation.
\end{itemize}

\section{Visualizations} \label{appendix:visualizations}

In this section, we show various qualitative comparisons and visualization of our method. 

\begin{figure}
\centering
\includegraphics[width=\linewidth]{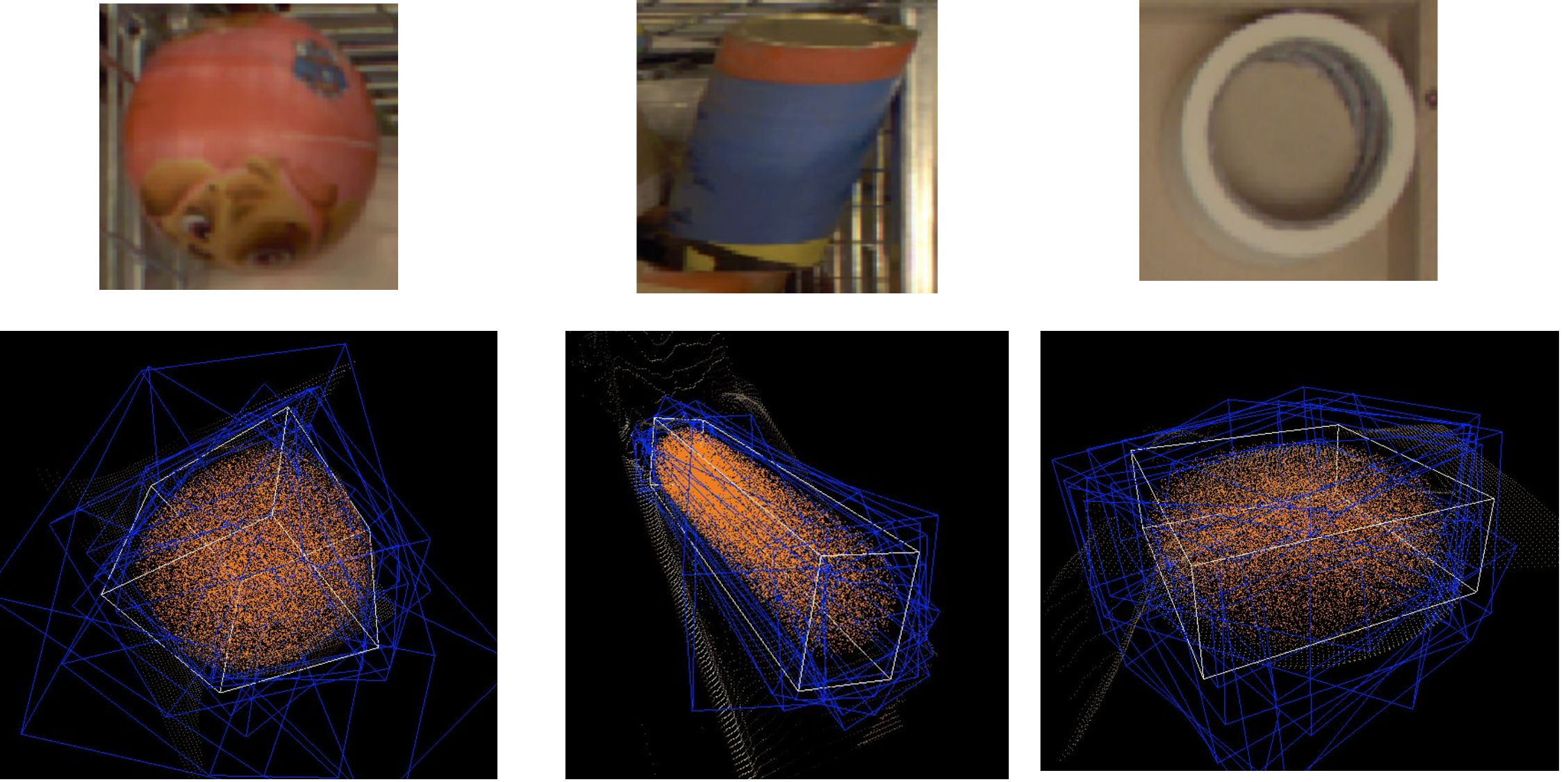}
\caption{Visualization of our model predictions on objects with rotational symmetry. The blue boxes show various samples from our model. The orange point cloud is the occupancy quantile. The white box is the quantile box. }
\label{fig:interesting-rotations}
\end{figure}

\begin{figure}
\centering
\includegraphics[width=\linewidth]{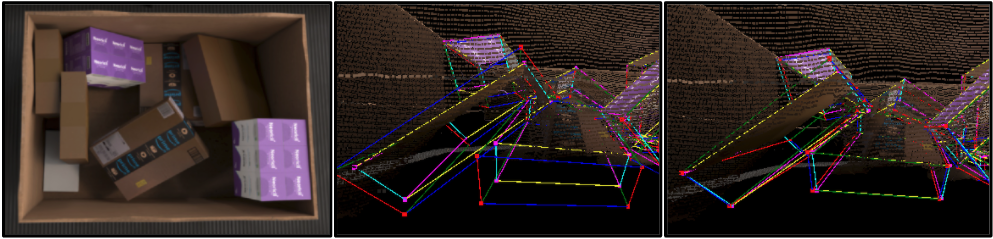}
\caption{Visualization of our dimension conditioning method. The model is able to leverage the conditioning information to accurately predict the correct pose \& dimension for each object's 3D bounding box. The prediction is shown in red-blue-green and the ground truth in turquoise-yellow-pink. Left: image of the scene. Middle: vanilla beam search. Right: beam search with dimension conditioning.}
\label{fig:multiple-conditioning-boxes}
\end{figure}

\begin{figure}
\centering
\includegraphics[width=\linewidth]{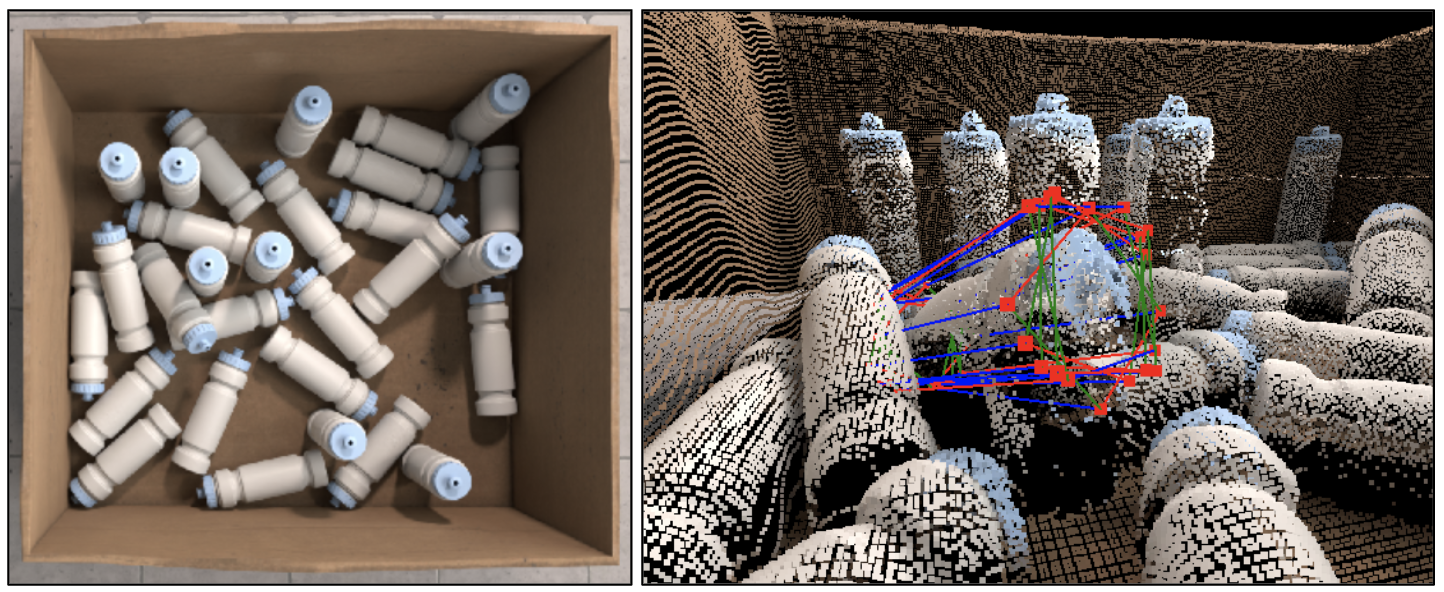}
\caption{Visualization of bounding box samples from our autoregressive model on a rotationally symmetric water bottle. Our model is able to sample different modes for symmetric objects whereas a deterministic model would only be able to predict a single mode.}
\end{figure}

\begin{figure}[t]
\centering
\includegraphics[width=\linewidth]{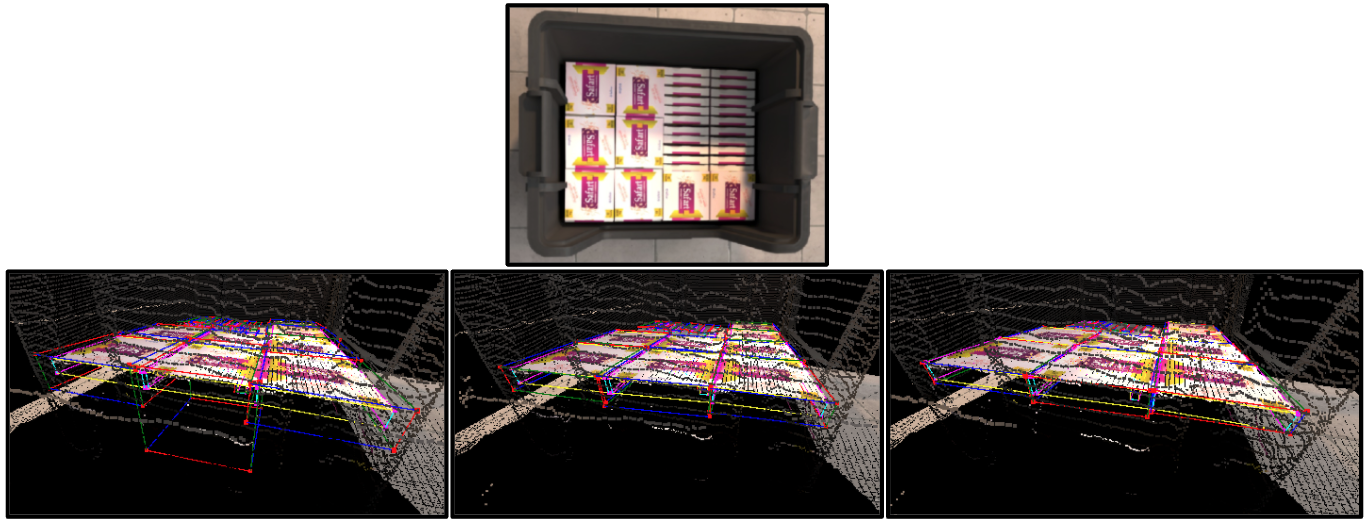}
\caption{Visualization of bounding box predictions with different quantiles. We can see that lower quantiles lead to larger boxes in the direction of uncertainty. Top: image of the scene. Left: quantile 0.1 Middle: quantile 0.3. Right: quantile 0.5.}
\end{figure}

\end{document}